\documentclass[10pt,twocolumn,letterpaper]{article}

\usepackage{iccv}
\usepackage{times}
\usepackage{epsfig}
\usepackage{graphicx}
\usepackage{amsmath}
\usepackage{amssymb}
\usepackage{authblk}

\makeatletter
\newcommand{\printfnsymbol}[1]{%
  \textsuperscript{\@fnsymbol{#1}}%
}
\makeatother

\usepackage[pagebackref=true,breaklinks=true,letterpaper=true,colorlinks,bookmarks=false]{hyperref}

\iccvfinalcopy 

\ificcvfinal\fi
\begin{document}

\title{OVSNet : Towards One-Pass Real-Time Video Object Segmentation}

\author{
Peng Sun$^{\dag}$\thanks{equal contribution}~~~~~
Peiwen Lin$^{\ddagger}$\printfnsymbol{1}~~~~~
Guangliang Cheng$^{\ddagger}$~~~~~
Jianping Shi$^{\ddagger}$~~~~~
Jiawan Zhang$^{\flat}$~~~~~
Xi Li$^{\dag}$
\\
\and
$^{\dag}$Zhejiang University~~~~$^{\ddagger}$SenseTime Research~~~~ $^{\flat}$Tianjin University\\
\and
{\tt\small 
\{sunpeng1996, xilizju\}@zju.edu.cn~~~jwzhang@tju.edu.cn}

{\tt\small 
\{linpeiwen, chengguangliang, shijianping\}@sensetime.com}
}


\maketitle

\begin{abstract}
Video object segmentation aims at accurately segmenting the target object regions across consecutive frames. It is technically challenging for coping with complicated factors
(e.g., shape deformations,  occlusion and out of the lens). Recent approaches have largely solved them by using back-forth re-identification and bi-directional mask propagation. However, their methods are extremely slow and only support offline inference, which in principle cannot be applied in real time. Motivated by this observation, we propose a efficient detection-based paradigm for video object segmentation. We propose an unified One-Pass Video Segmentation framework (OVS-Net) for modeling spatial-temporal representation in a unified pipeline, which seamlessly integrates object detection, object segmentation, and object re-identification. The proposed framework lends itself to one-pass inference that effectively and efficiently performs video object segmentation. Moreover, we propose a mask-guided attention module for modeling the multi-scale object boundary and multi-level feature fusion. Experiments on the challenging DAVIS 2017 demonstrate the effectiveness of the proposed framework with comparable performance to the state-of-the-art, and the great efficiency about \textbf{11.5 FPS} towards pioneering real-time work to our knowledge, more than \textbf{5 times} faster than other state-of-the-art methods.
\end{abstract}

\begin{figure}[t]
\centering
\includegraphics[width=8cm]{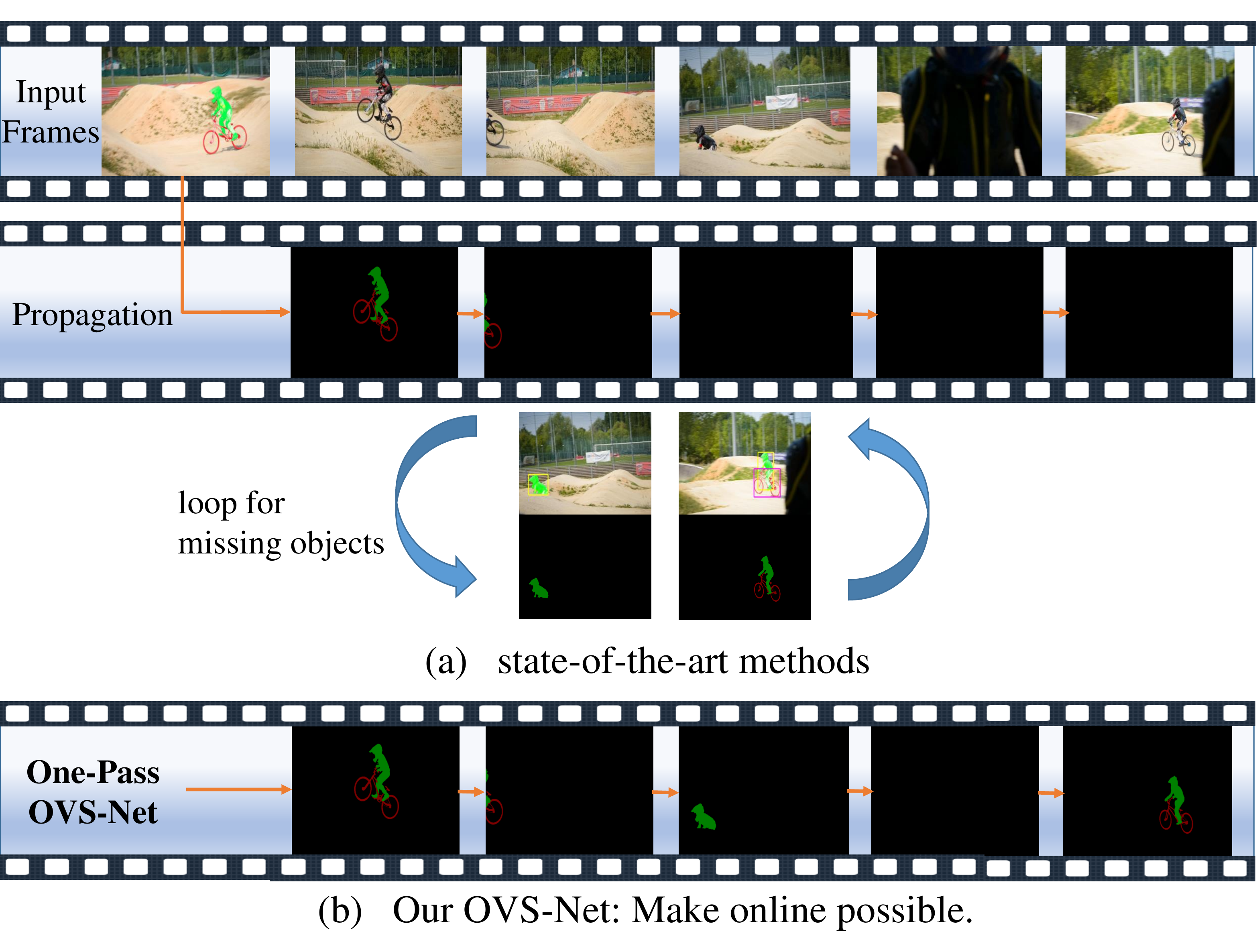}
\caption{We focus on the person and the bicycle in the original input frames.~As shown in (a), the state-of-the-art methods such as VS-ReID or DyeNet need traverse the video many times to retrieve the missing objects due to occlusion or out of the lens. {Our OVS-Net can produce accurate segmentation results in a one-pass manner, as shown in (b).}   {Best viewed in color.}}

\label{fig:flowchart}
\end{figure}

\section{Introduction}

As an important and challenging problem in computer
vision, video object segmentation is typically cast as a problem of instance-aware pixel-wise classification across consecutive frames with the object mask provided in the first
frame. In principle, video object segmentation requires accomplishing the following two tasks: 1) intra-frame object
instance segmentation; and 2) inter-frame object instance
association. With the single first frame label, how to effectively model spatio-temporal context is key issue for robust video object segmentation.

Traditional methods~\cite{conf/iccv/LeeKG11, conf/iccv/PapazoglouF13, conf/eccv/XuXC12, conf/cvpr/GrundmannKHE10a,conf/cvpr/HicksonBEC14} are graph based to build spatio-temporal graphs for contextual relationship in video object segmentation.~With the power of deep learning, such spatial-temporal correlation is usually carried out within a deep neural network framework, which takes an end-to-end convolution neural network learning architecture over the benchmark datasets (e.g., DAVIS Challenge~\cite{davis2017,davis2016}).~One stream of spatial-temporal modeling is via optical flow across frames \cite{conf/cvpr/Tsai0B16,khoreva2017lucid,conf/cvpr/JampaniGG17}.~Such optical flow based methods \cite{khoreva2017lucid,conf/cvpr/Tsai0B16,conf/cvpr/JampaniGG17} are prone to errors at occluded or reappearing objects, which commonly appears in real world.~In order to address the problem of object occlusion, vanishing, or reappearing, another stream of methods \cite{xiaoxiao-18video,xiaoxiao2017video,conf/cvprw/IRIF} utilize object re-identification (Re-ID)  techniques \cite{conf/cvpr/XiaoLWLW17} where the occlusion or reappearing problem is largely solved by the powerful Re-ID features.~However, as shown in Figure \ref{fig:flowchart}(a), the aforementioned approaches \cite{xiaoxiao-18video,xiaoxiao2017video} essentially adopt an offline Re-ID strategy (back-forth Re-ID) that traverses the whole videos for several times and then performs bi-directional mask propagation during inference, which is far away from the real-time or online requirements.~

To enjoy the ability to handle occlusion or reappearing problems and towards real-time application, we propose to build a detection-based pipeline as well as an effective unified framework named OVS-Net for modeling spatial-temporal representation.~The proposed framework takes a seamless integration of object detection, object segmentation, and object re-identification.~Specifically, the proposed framework carries out the joint learning procedure of object segmentation and object Re-ID during training.

To enable efficient inference and associate the corresponding outputs by OVS-Net across frames, we propose an one-pass strategy to produce accurate segmentation results frame by frame in a one-pass manner with online Re-ID, as shown in Figure \ref{fig:flowchart}(b). The one-pass strategy is based on a cascaded rule consisting of three paths: IOU path,
Re-ID path, and flow path, resulting in great computational efficiency.~Furthermore, we propose a mask-guided attention module to fully explore
the multi-scale object information and multi-level feature fusion, which leads to performance improvements in object boundary.~As a matter of fact, the joint training aims at enabling the network to learn the capability of object segmentation within individual frames and object Re-ID association across different frames. In contrast, the motivation of the cascaded inference strategy is to ensure object segmentation towards the real-time performance.

Specifically, the instance masks generated in~\cite{xiaoxiao-18video, xiaoxiao2017video} are warped from previous frames using optical flow and then refined to get better predictions. 
Instead, our method generates masks independently for each frame and then links these masks through the proposed one-pass cascaded inference strategy.~The pipeline we propose can be seen as a new unified detection-based paradigm different against their propagation-based methods~\cite{xiaoxiao-18video,xiaoxiao2017video}.~Our approach achieves a comparable performance to the state-of-the-art approaches and can perform  \emph{11.5 frames per seconds}.~Our OVS-Net is about \emph{5 times} faster than DyeNet \cite{xiaoxiao-18video}, \emph{35 times} faster than VS-ReID \cite{xiaoxiao2017video}.~Extensive experiments demonstrate the performance effectiveness and efficiency of this work above
against the state-of-the-art.

The main contributions of this work can be
summarized as follows.

\begin{itemize}
	\item We propose the first unified detection-based pipeline for video object segmentation as well as a joint training framework for
	modeling spatio-temporal representation, which seamlessly integrates the modules of object detection, segmentation, and re-identification in a unified manner.
	
	\item We propose an effective one-pass cascaded inference strategy, which leads the network inference  towards real-time performance (around \emph{11.5 fps}) in a one-pass manner with online Re-ID.
	
	\item Furthermore, we propose a mask-guided attention module to fully explore the multi-scale object information within a feature pyramid network, which leads to performance improvements in object boundary segmentation.
	
\end{itemize}

\section{Related Work}

\par \textbf{Semantic Segmentation}~~Video object segmentation pays great attention to the quality of segmentation results.~With the boom of deep learning, semantic segmentation~\cite{journals/pami/ChenPKMY18, journals/pami/LiuLLLT18,pspnet,deeplabv3, conf/iccv/0001JRVSDHT15} have shown extremely good performance in static images.~Different from semantic segmentation which has pre-defined semantic label, in video object segmentation, we focus on the class-agnostic object segmenation.

\par \textbf{Instance Segmentation}~~In video object segmentation, we also need to distinguish between different instances.~There are many effective methods in instance segmentation \cite{conf/eccv/HariharanAGM14, conf/cvpr/HariharanAGM15,he2017maskrcnn, conf/cvpr/LiQDJW17, journals/corr/abs-1803-01534}, and Mask R-CNN is the most popular solution recently.~Our framework is built on Mask R-CNN and improves it from many aspects, which is the first effective unified detection-based framework for video object segmentation.

\begin{figure*}
\begin{center}
\includegraphics[width=7in]{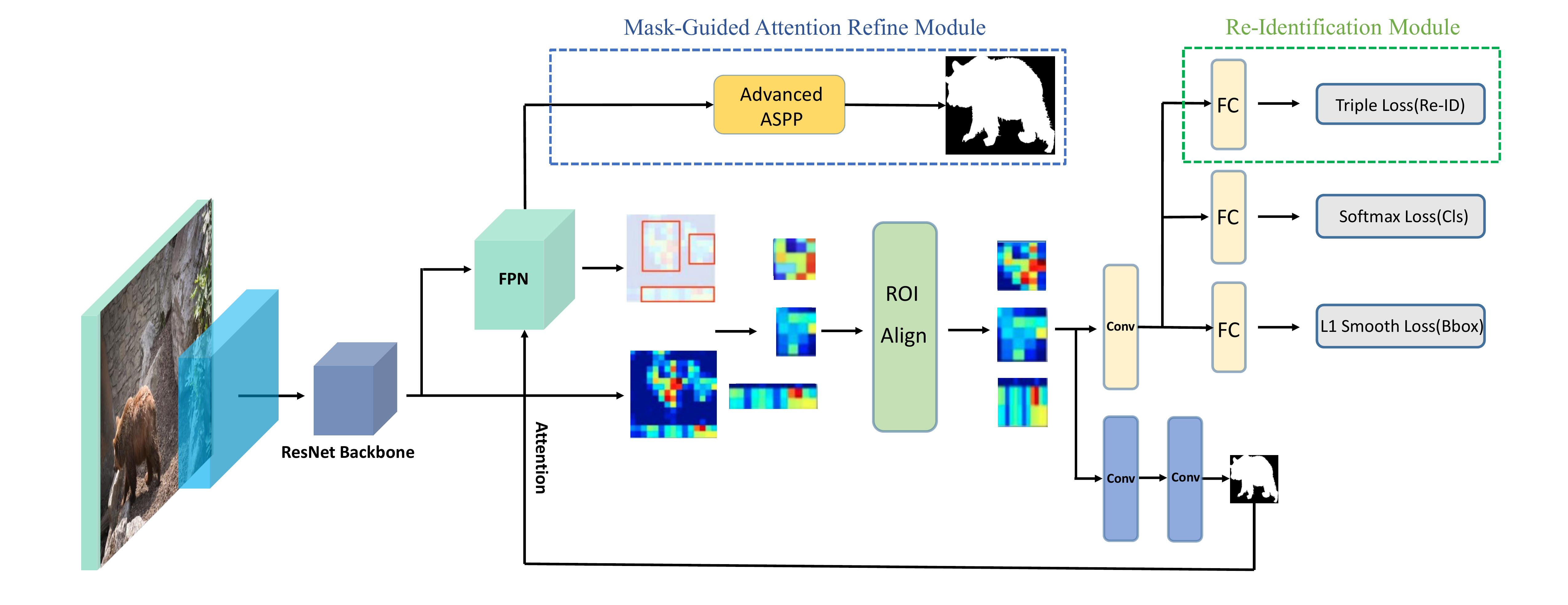}
\end{center}
   \caption{Illustration of our framework. The Re-ID branch is our online Re-Identification module and the middle part of the above is the Mask-guided Attention module which use the Mask R-CNN prediction as the attention map and combine the muti-level features of FPN for producing more accuracy segmentation result than original Mask R-CNN. }
\label{fig:framework}
\end{figure*}


\par \textbf{Video Object Segmentation}~~Recently, deep learning based model is the most promising one to tackle video object segmentation task. For example, Perazzi et al.~\cite{conf/cvpr/PerazziKBSS17} introduced video object segmentation as a guided instance segmentation problem via frame-by-frame pixel-labeling strategy. Caelles et al.~\cite{OSVOS} introduced one transfer learning based algorithm (OSVOS) to tackle the task of foreground segmentation. While OSVOS can achieve impressive performance, it is slightly sensitive to large changes in object appearance. To tackle this limitation, Voigtlaender et al.~\cite{conf/bmvc/VoigtlaenderL17} proposed an online adaptive video object segmentation algorithm that can update the network online using the selected training examples. Tsai et al.~\cite{conf/cvpr/Tsai0B16} considered the video segmentation through a optical flow based algorithm to maintain object boundaries and temporal consistency simultaneously. Li et al.~\cite{conf/iccv/0003ZCZ17} incorporated the neighborhood reversible flow to segment the foreground objects and suppress the distractors. Video propagation networks~\cite{conf/cvpr/JampaniGG17} incorporated temporal bilateral network and adaptive filtering strategy to propagate information forward to the future frames.~However, they are unable to track the object that re-appears in the video.

To tackle the above limitation, some state-of-the-art approaches~\cite{conf/cvprw/IRIF,xiaoxiao-18video,xiaoxiao2017video} incorporated person re-identification algorithm~\cite{conf/cvpr/XiaoLWLW17} into the video object segmentation framework.~For instance, Instance Re-Identification Flow (IRIF)~\cite{conf/cvprw/IRIF} can track and detect the re-appeared instance via the instance re-identification module and mask propagation module. Li et al.~\cite{xiaoxiao2017video} adapted Re-ID approach and a two-stream mask propagation model in their framework, while it is slow in inference and slightly sensitive to the pose variations. To overcome these shortcomings, Li et al.~\cite{xiaoxiao-18video} proposed a substantially robust and efficient network with the attention mechanism, while their back-forth Re-ID and bi-directional propagation methods are offline and takes about 0.43 seconds per frame, which is far from real-time application.~Meanwhile, different from their propagation-based methods \cite{xiaoxiao-18video,xiaoxiao2017video}, we propose the first unified detection-based pipeline, in this way, we do not need to use optical flow information during the training process, which simplifies the training process.~In addition, the attention mechanism proposed in~\cite{xiaoxiao-18video} is in propagation process, while our attention module is for better modeling the intra-frame spatial context information.~Furthermore, although \cite{xiaoxiao-18video} conducted one-iteration Re-ID experiment, they still adopted an offline back-forth greedy strategy (offline sorted for Re-ID and bi-directional propagation), which cannot be used to deal with online or streaming videos.~Though our proposed unified detection-based pipeline, we can perform online Re-ID and achieve comparable performance, meanwhile our OVS-Net is about \emph{5 times} faster than DyeNet \cite{xiaoxiao-18video}.

Another similar work to us is PReMVOS~\cite{premvos2018}. They proposed a proposal merging algorithm and achieved state-of-the-art results across all datasets. Different from their using many networks in~\cite{premvos2018} such as Mask R-CNN network, Proposal-Refinement Network, Re-ID embedding network and so on, we propose a new brief unified training framework which only contains one network.~Furthermore, they proposed a greedy proposal merging algorithm, and ran the random-search hyper-parameter optimisation for 25000 random parameter values. They also adopted an offline greedy strategy, which cannot be used to deal with online or streaming videos.

\section{Framework}

In this section, we first introduce our effective detection-based unified framework during training phase, for intra-frame spatial context modeling and inter-frame temporal object association in an unified scheme by describing two sub-modules: Mask-Guided Attention (MGA) module and Re-ID module.~Later we describe our one-pass cascaded strategy to perform object segmentation within individual frames and sequentially link the corresponding ID-specific object segmentation masks frame by frame in a one-pass manner towards real-time.

\begin{figure*}
\begin{center}
\includegraphics[width=7in]{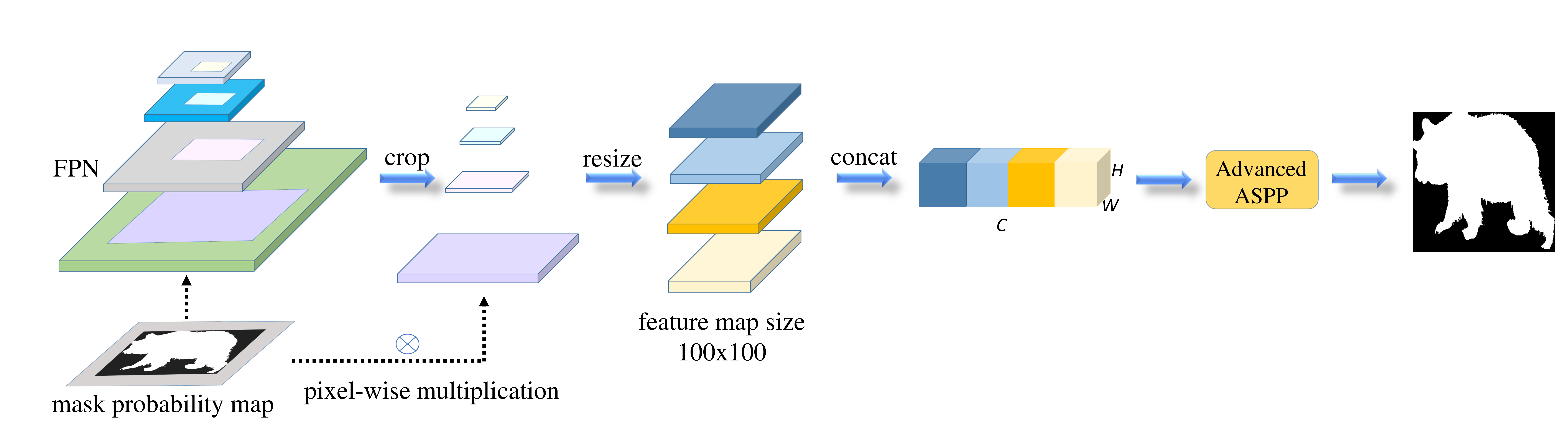}
\end{center}
   \caption{ Illustration of the Mask-Guided Attention Module.}
\label{fig:refine_module}
\end{figure*}

\subsection{Fully Unified Network}

Given a video and the target object masks in the first frame, our goal is to segment the target instance objects in future frames.~The proposed unified framework seamlessly integrates the object detection, object segmentation and object re-identification.~Specifically, it is the first unified detection-based pipeline for video object segmentation.~As illustrated in Figure~\ref{fig:framework}, our framework consists of Mask R-CNN~\cite{he2017maskrcnn}, online Re-Identification (Re-ID) module and MGA module. Specifically, the MGA module utilizes an attention strategy to fuse multi-level information that can greatly improve the object boundary performance, for better intra-frame spatial context modeling.~The Re-ID module is capable of retrieving the reappeared objects and is incorporated into the Mask R-CNN~\cite{he2017maskrcnn} network, which can adapt to each other during the joint optimization,~for inter-frame temporal object association.
\vspace{6pt}


\subsubsection{Mask-Guided Attention Module}

\indent \par First, we model the intra-frame spatial context information for each image. We use Mask R-CNN~\cite{he2017maskrcnn} as our base instance segmentation modeling method. Generally, the output size of Mask R-CNN is $28 \times 28$, which is relatively small that can not achieve satisfactory aligned boundary well, especially for big objects.~To fully explore the multi-scale object information, a powerful module named Mask-Guided Attention (MGA) is proposed, which use the prediction of Mask R-CNN as the attention map and combine the muti-level features of FPN~\cite{fpn} to produce bigger and more accurate segmentation result than the original output of Mask R-CNN without introducing a lot of time overhead.

Different from original Mask R-CNN head, here we use the MGA module to predict a class-agnostic foreground/background mask. In addition, the MGA module will expand the attention masks and features at the corresponding position by 20\% (for the attention mask, we fill it with a fixed probability value of 0.5) in order to segment the boundary of the object accurately. We use the probability map of the output of Mask R-CNN as the attention map, crop the feature maps of FPN according to the coordinate of enlarged bounding bboxes, and pixel-wise multiply the cropped feature maps with the attention map, then resize the feature maps to 100x100 and pass the last block of ResNet-101~\cite{resnet} followed by the advanced ASPP module to get the final segmentation result. The loss function is the pixel-wise cross entropy loss and the gradient generated by MGA module will only propagate backward to itself, and will not affect FPN and Mask R-CNN. Figure \ref{fig:refine_module} illustrates the Mask-guided Attention module. For the Advanced ASPP module, we simply combine the ASPP module of DeepLabv3~\cite{deeplabv3} and the PPM module of PSPNet~\cite{pspnet}, which bring a slight improvement in performance.

\subsubsection{Online Re-identification Module}

\indent \par Since we have modeled the intra-frame spatial context information, now we focus on inter-frame object instance association.~
We propose the powerful online re-identification module to find the reappeared objects in real time. In all experiments, we assign a unique, non-repeating re-identification label for each target object in each video. We use the Triplet-Loss~\cite{facenet} for training the online re-identification module, we project the features extracted by RoI-Align Pooling into a 128-dimensional feature vector by simply adding two fully connected (fc) layers and the last {\em fc} layer is followed by a dropout layer with a dropout rate 0.2.~In principle, our training framework adopts a multi-task learning pipeline that jointly optimizes the Re-ID module in conjunction with the Mask-RCNN module on the basis of the same backbone network.~The joint training process is carried out for each minibatch consisting of different frames from the same or different videos. For the Re-ID training, we select N bbox proposals in each frame, and then get its Re-ID feature through the Re-ID module.~Suppose each minibatch has M frames, as a result, we totally have M*N training examples from diverse (the same or different) frames, enough to use the triplet loss.~The Re-ID training aims to learn object association for the between-frame training examples,
while the Mask-RCNN training seeks for learning the spatial segmentation for the within-frame training examples.
Finally, the learning gradients of the above two modules are simultaneously back-propagated to the shared
backbone network for the joint optimization process.
\par The loss function of the online Re-ID module is as follows,

\begin{small}
\begin{equation}
L_{reid}  \! =  \! \sum_{i=1}^{N} \left [ \left \| f(x_{i}^{a}) - f(x_{i}^{p}) \right \|_{2}^{2} -  \left \| f(x_{i}^{a}) - f(x_{i}^{n}) \right \|_{2}^{2}  + \alpha  \right ]_{+},
\end{equation}
\end{small}where {\em p} denotes positive object set that has the same Re-ID label with object {\em a}, and $n$ denotes positive object set that has different Re-ID labels with object {\em a}. {\em f(x$_{i}^{a}$)} is the reid-feature for object {\em a}, and {\em f(x$_{i}^{p}$)} is the reid-feature for the positive object that is the most dissimilar to the object {\em a}, and {\em f(x$_{i}^{n}$)} is the reid-feature for the negative object that is most similar to the object {\em a}. {\em $\alpha$} is the margin of triplet-loss and we experimentally set it to 1.0.

\subsection{One-Pass Cascaded Inference Strategy}

\par
\begin{figure}[t]
\centering
\includegraphics[width=3.5in]{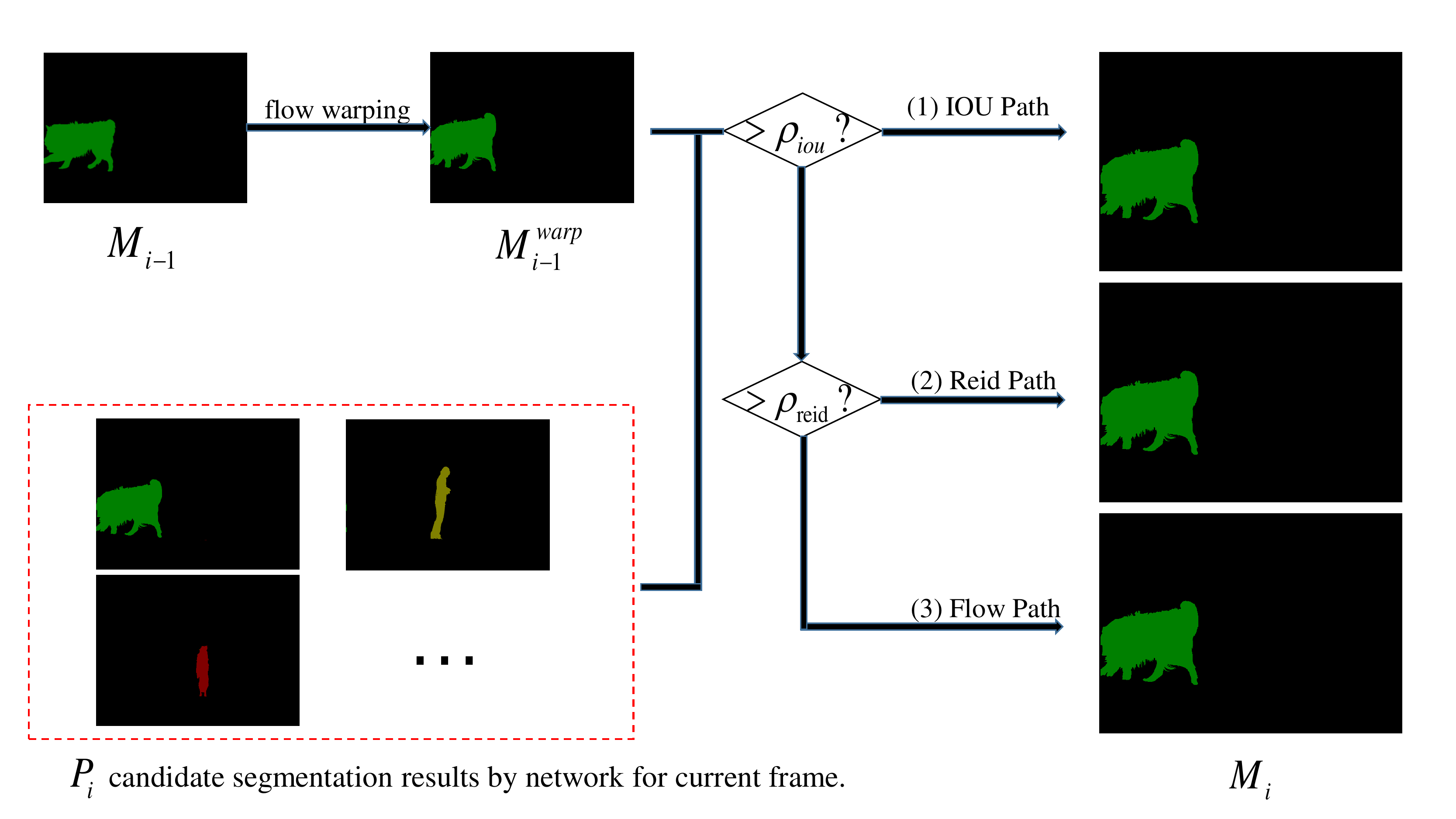}
\caption{ The process of our one-pass cascaded inference strategy. \textbf{Best viewed in color.}}
\label{fig:inference}
\end{figure}

 With our well-designed inference strategy, the video frames need to traverse only once rather than many times, which makes it possible for the video object segmentation in real time.~In principle, the cascaded inference strategy makes full use of smooth transition across adjacent frames to
locate the object segmentation results by lightweight optical flow, while long-term inter-frame objects after occlusion or drastic motion can be handled by Re-ID.~The object association rules are based on a cascaded strategy consisting of three paths: IOU path, Re-ID path and flow path.~We define $\rho$ $_{reid}$, $\rho$ $_{iou}$, {\em C$_{reid}^{n}$} as the similarity threshold for reid-feature, the threshold for the IOU and the reid-feature collection for {\em n-th} instance, respectively.~So the number of collection {\em C$_{reid}$} is equal to the number of instances need to track in the video.~As shown in Figure \ref{fig:inference}, given the mask of last frame {\em M$_{i-1}$} and the candidate objects {\em P$_{i}$} predicted by our OVS-Net, we aim to generate the probability map for certain instance in current frame {\em M$_{i}$} by three cascaded inference paths, which is IOU-Path, Reid-Path, and Flow-Path.

To better explain above three paths, we choose the video with only one instance as our example and denote this instance as ${I}$. If there are multiple instances in the video, the same operation will be simply executed for all instances.~Before executing these three operation, we first warp the mask of last frame for instance ${I}$ by optical flow extracted by FlowNet 2.0~\cite{conf/cvpr/IlgMSKDB17} as the mask template {\em M$_{i-1}^{warp}$} for current frame and then generate the candidate segmentation result {\em P$_{i}$} and its reid-feature in current frame by our OVS-Net. After that, the following three operation will be executed:

\textbf{IOU-Path}: As mentioned above, we have obtained the candidate segmentation results for current frame and its reid-features. In this operation, we first calculate the IOU between all candidates and the instance ${I}$ in the mask template {\em M$_{i-1}^{warp}$}. If there exist candidates whose IOU are larger than $\rho$$_{iou}$, we take the candidate that has the largest IOU as the final mask for the instance ${I}$ in current frame and also save its reid-feature in the instance ${I}$'s reid-feature collection {\em C$_{reid}^{I}$}, otherwise, the program will execute the next operation Reid-Path.

\textbf{Reid-Path}: When the IOU-Path setting can't be met, it is possible that we have lost instance ${I}$ in previous frame, so we need to use ${I}$'s reid-feature saved in C$_{reid}^{I}$ to retrieve instance ${I}$ in current frame. Firstly, The similarity between all candidate reid-features and the instance ${I}$'s reid-features saved in {\em C$_{reid}^{I}$} will be calculated through measuring the euclidean distance. If there exist candidates whose similarity with at least $30\%$ reid-features in collection {\em C$_{reid}^{I}$} is smaller than $\rho$$_{reid}$, we take the candidate who has the smallest similarity with instance ${I}$ as its final mask in current frame.

\textbf{Flow-Path}: If no any candidates satisfy the paths above, a rough mask warped by optical flow {\em M$_{i-1}^{warp}$} will be sent to MGA module to generate the final mask for instance ${I}$ in current frame.

\section{Experiments}

\subsection{Implementation Details }

We re-implement Mask R-CNN and FPN based on Pytorch 0.3.1~\cite{paszke2017automatic} and use ResNet-50~\cite{resnet} as our backbone. We train the entire network starting from pre-trained ImageNet~\cite{deng2009imagenet} weights on COCO~\cite{coco} dataset. Follow the common configurations, we use the muti-scale training, the shorter edges of the input images are 640, 720, 800, 920, 1080, the max edge of the images is 1900 and the anchor scales are 3, 5, 8.
~After pretrained on COCO, the origin Mask R-CNN is first trained on DAVIS training sets for two epochs in order to initialize the whole Network. Then the whole network with Re-ID module and the Mask-guided Attention module is trained on DAVIS for three epochs. We fix a mini-batch size of 32 images for Mask R-CNN, momentum 0.9 and weight decay of 10$^{-4}$, and for every image in a mini-batch, we select two proposals which have the largest IoU with ground-truth bboxes for the MGA Module and Re-ID module. So the mini-batch of the Re-ID module and the MGA Module are both 64. The initial learning is 0.04 and dropped by a factor of 10 after each epoch. For data augmentation, we employ the Lucid Data Dreaming~\cite{khoreva2017lucid} to generate augmented images using the first frame on testing videos and add them into the training set that adopts our model to the target video domain. The overall loss function of our model is formulated as:

\begin{equation}
L_{total} = L_{mask}+ L_{loc}+ L_{cls}+ L_{reid}+ \lambda L_{refine}
\end{equation}

The {\em L$_{reid}$} is the triplet loss of Re-ID module. {\em L$_{loc}$}, {\em L$_{cls}$}, {\em L$_{mask}$} respectively represent location loss, classification loss and segmentation loss of the Mask R-CNN. The {\em L$_{refine}$} is the pixel-wise cross-entropy loss of the MGA module where {\em $\lambda$} is a weight that balances these loss terms for better muti-task learning. During inference phase, we only use the single scale without muti-scale and horizontal flip testing and we keep the proposals with a score greater than 0.05 and perform non-maximum suppression which have an IOU of 0.6 with a proposal with a higher score.

\begin{figure*}
\begin{center}
\includegraphics[width=7in]{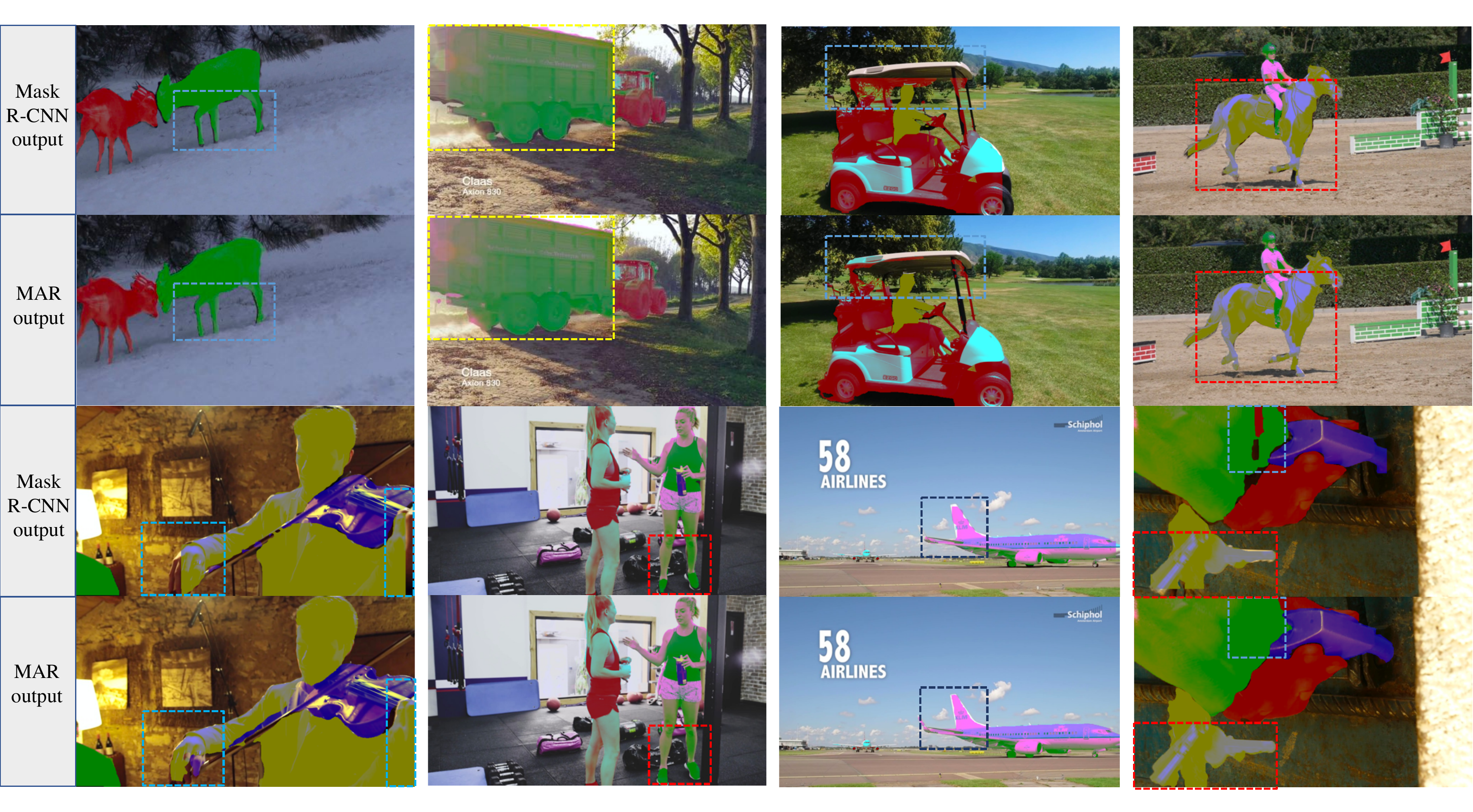}
\end{center}
   \caption{Visualization results with or without the Mask-guided Attention Module. The following line is the result with the Mask-guided Attention Module, having better segmentation result.~The dashed box is drawn for easy comparison. \textbf{Best viewed in color.}}
\label{fig:compare}
\end{figure*}

\subsection{Benchmark}

\textbf{Datasets}  In order to verify the effectiveness and robustness of our method, we evaluate our method on DAVIS$_{16}$~\cite{davis2016}, DAVIS$_{17}$~\cite{davis2017} and SegTrack$_{v2}$~\cite{segtrackv2} datasets. The DAVIS dataset is a public dataset, benchmark, and competition specially designed for the task of video object segmentation, spanning multiple occurrences of common video object segmentation challenges such as occlusions, fast-motion, appearance changes and out of the lens.~SegTrack$_{v2}$ dataset is a video segmentation dataset with full pixel-level annotations on multiple objects in each frame within each video. We conduct a complete ablation study on the DAVIS$_{17}$ test-dev dataset. In this section, we compare the first unified detection-based one-pass OVS-Net with other existing state-of-the-art methods and show it can achieve the comparable performance towards real-time on these standard datasets. 

\textbf{Evaluation Metric} For DAVIS$_{17}$ Dataset, we follow~\cite{davis2017} that employs the Jaccard index {\em J} defined as the {\em intersection-over-union} of the estimated segmentation and the groundtruth mask and employ Contour Accuracy {\em F} to compute the contour-based precision and recall and their average {\em G} measures for evaluation. For DAVIS$_{16}$ and SegTrack$_{v2}$ Datasets, we use the Jaccard index {\em J} defined as the {\em intersection-over-union} across all instances to evaluate the performance, same as other methods.

\textbf{Different testing patterns} Following \cite{premvos2018}, training
modalities can be divided into \textbf{Only-Training-Set training} (OTS) and \textbf{Training-Testing-Set training} (TTS). In only-train-set training a model is only trained on the training set without any annotations from the test set as described in~\cite{xiaoxiao-18video}. Since the first frame annotations are provided in the testing stage, we can use them for finetuning the model, namely train-test-set training. For testing phase, there are two patterns, they can be further divided into \textbf{per-dataset} and \textbf{per-video} finetuning. In per-dataset finetuning, we merge all first frame annotations from test set into training set that adopts the model to the target video domain to obtain a dataset-specific model. However, per-video finetuing means that finetune a model on each testing video, i.e., the number of final models is as many as the number of videos during the test phase. Obviously, the former has better universality. Table \ref{davis17} lists the {\em J}-means, {\em F}-means and {\em G}-means on DAVIS$_{17}$ {\em test-dev} dataset with other existing state-of-the-art methods. And for DAVIS$_{16}$, we use its own train set and its  first frames in val set to train our model and evaluate the model on DAVIS$_{16}$ val set. For SegTrack$_{v2}$, we only use the first frames of the videos as the training data and finetune the model which is pretrained on DAVIS$_{17}$. And Table \ref{davis16} lists the performance on DAVIS$_{16}$ val set and SegTrack$_{v2}$ dataset. Due to memory limitation, the backbone used in our all experiments is ResNet50~\cite{resnet}.


\begin{table}[h]
\begin{center}
\setlength{\tabcolsep}{1mm}{
\begin{tabular}{|l|c|c|c|c|}
\hline
Method & {\em J}-mean & {\em F}-mean & {\em G}-mean & per-video \\
\hline\hline
OSVOS~\cite{OSVOS} & 47.0 & 54.8 & 50.9 & \checkmark \\
OSVOS-S~\cite{OSVOS-S} & 52.9 & 62.1 & 57.5  & \checkmark \\
OSAVOS~\cite{conf/bmvc/VoigtlaenderL17} & 53.4 & 59.6 & 56.5 & \checkmark\\
LucidTracker$^{*}$~\cite{khoreva2017lucid} & 60.1 & 68.3 & 64.2 & \checkmark\\
DyeNet~\cite{xiaoxiao-18video} & 65.8 & 70.5 & 68.2 & $\times$ \\
DyeNet (OTS)~\cite{xiaoxiao-18video}& 60.2 & 64.8 & 62.5 & $\times$ \\
PReMVOS \cite{premvos2018} & 67.5 & 76.8 & 71.6 & $\times$ \\
VS-ReID$^{*}$~\cite{xiaoxiao2017video} &  63.3 & 67.0 & 65.2 & $\times$ \\
\hline
OVS-Net$^*$& 62.5 & 68.4 & 65.5 & $\times$ \\
OVS-Net (OTS) & 59.3 & 64.6 & 62.0 & $\times$ \\
\hline
\end{tabular}}
\end{center}
\caption{Results on DAVIS$_{17}$ test-dev dataset. * means only using the single scale for testing and per-video means whether to use the per-video finetuning..}
\label{davis17}
\end{table}

\begin{table}[h]
\begin{center}
\setlength{\tabcolsep}{1mm}{
\begin{tabular}{|l|c|c|c|}
\hline
Method & Davis$_{16}$ & SegTrack$_{v2}$  & per-video\\
\hline\hline
OSVOS~\cite{OSVOS} & 79.8 & 65.4 & \checkmark  \\
MSK~\cite{conf/cvpr/PerazziKBSS17} & 80.3 & 70.3 & \checkmark \\
OSAVOS~\cite{conf/bmvc/VoigtlaenderL17} & 85.7 & - & \checkmark \\
LucidTracker~\cite{khoreva2017lucid} &  84.8 & 77.6 & \checkmark  \\
DyeNet~\cite{xiaoxiao-18video}& 84.7 & 78.7 & $\times$  \\
PReMVOS \cite{premvos2018} & 86.8 & - &  \\
\hline
OVS-Net & 84.6 & 76.9 & $\times$ \\
\hline
\end{tabular}}
\end{center}
\caption{Results on DAVIS$_{16}$ dataset and SegTrackv2 datasets. Per-video means whether to use the per-video finetuning.}
\label{davis16}
\end{table}

\subsection{Ablation Study}

In this section, in order to verify the effectiveness of each component of our framework, we perform the ablation study experiments.~All performance are reported on the {\em test-dev} set of DAVIS$_{17}$. \\

\noindent \textbf{Robustness of the Online Re-Identification Module}

For Re-ID module, we performed a series of experiments with different $\rho$$_{reid}$ value in Re-ID path or without the Re-ID module. The $\rho$$_{reid}$  is extremely important for recall and precision. Table \ref{reid_ab} shows the different $\rho$$_{reid}$ values with their performance. And using the $\rho$$_{reid}$ = 2.3, our OVS-Net achieves the best performance. In the following experiments, we fix the $\rho$$_{reid}$ as 2.3. 

\begin{table}[h]
\begin{center}
\setlength{\tabcolsep}{0.8mm}{
\begin{tabular}{|c|c|c|c|}
\hline
Methods  & {\em J}-mean & {\em F}-mean & {\em G}-mean \\
\hline\hline
OVS-Net(without Re-ID) & 53.1 & 56.7 & 54.9  \\
OVS-Net(with $\rho$$_{reid}$ = 2.1) & 56.9 & 63.7 & 60.3  \\
OVS-Net(with $\rho$$_{reid}$ = 2.2) & 57.6  & 63.2 & 60.4  \\
OVS-Net(with $\rho$$_{reid}$ = 2.3) & \textbf{58.3} & \textbf{63.7} & \textbf{61.0}  \\
OVS-Net(with $\rho$$_{reid}$ = 2.4) & 57.4 & 63.2 & 60.3  \\
OVS-Net(with $\rho$$_{reid}$ = 2.5) & 57.0 & 63.0 & 60.0  \\
\hline
\end{tabular}}
\end{center}
\caption{Ablation study for Re-ID module with different $\rho$$_{reid}$ value on DAVIS$_{17}$ test-dev dataset.}
\label{reid_ab}
\end{table}

\noindent \textbf{Effectiveness of the Mask-guided Attention Module} 

The Mask-guided Attention Module is very effective, to prove this, we performed a series of ablation study experiments. Our network is essentially multi-task learning, and the loss weight between different tasks is very important in multi-task learning. Figure \ref{fig:compare} shows that the segmentation result of our Mask-guided Attention Module is better than the original Mask R-CNN. And Table \ref{MGARM_ab} lists the results with different loss weight $\lambda$. We can conclude that MGA module brings above 3 points performance than baseline OVS-Net. We find that the OVS-Net achieves the best performance when $\lambda$ is 1.3.

\begin{table}[h]
\begin{center}
\setlength{\tabcolsep}{1mm}{
\begin{tabular}{|c|c|c|c|}
\hline
Methods  & {\em J}-mean & {\em F}-mean & {\em G}-mean \\
\hline\hline
OVS-Net(without MGA) & 58.3 & 63.7 & 61.0 \\
OVS-Net(with $\lambda$ = 1.0) & 62.1 & 66.7 & 64.4  \\
OVS-Net(with $\lambda$ = 1.1) & 63.2 & 67.4 & 65.3  \\
OVS-Net(with $\lambda$ = 1.2) & 62.1 & 67.6 & 64.9  \\
OVS-Net(with $\lambda$ = 1.3) & \textbf{62.5} & \textbf{68.4} & \textbf{65.5} \\
OVS-Net(with $\lambda$ = 1.4) & 62.0 & 67.8 & 64.9 \\
\hline
\end{tabular}}
\end{center}
\caption{Ablation study for Mask-guided Attention Module with DAVIS$_{17}$ test-dev dataset.}
\label{MGARM_ab}
\end{table}

\begin{figure*}
\begin{center}
\includegraphics[width=7in]{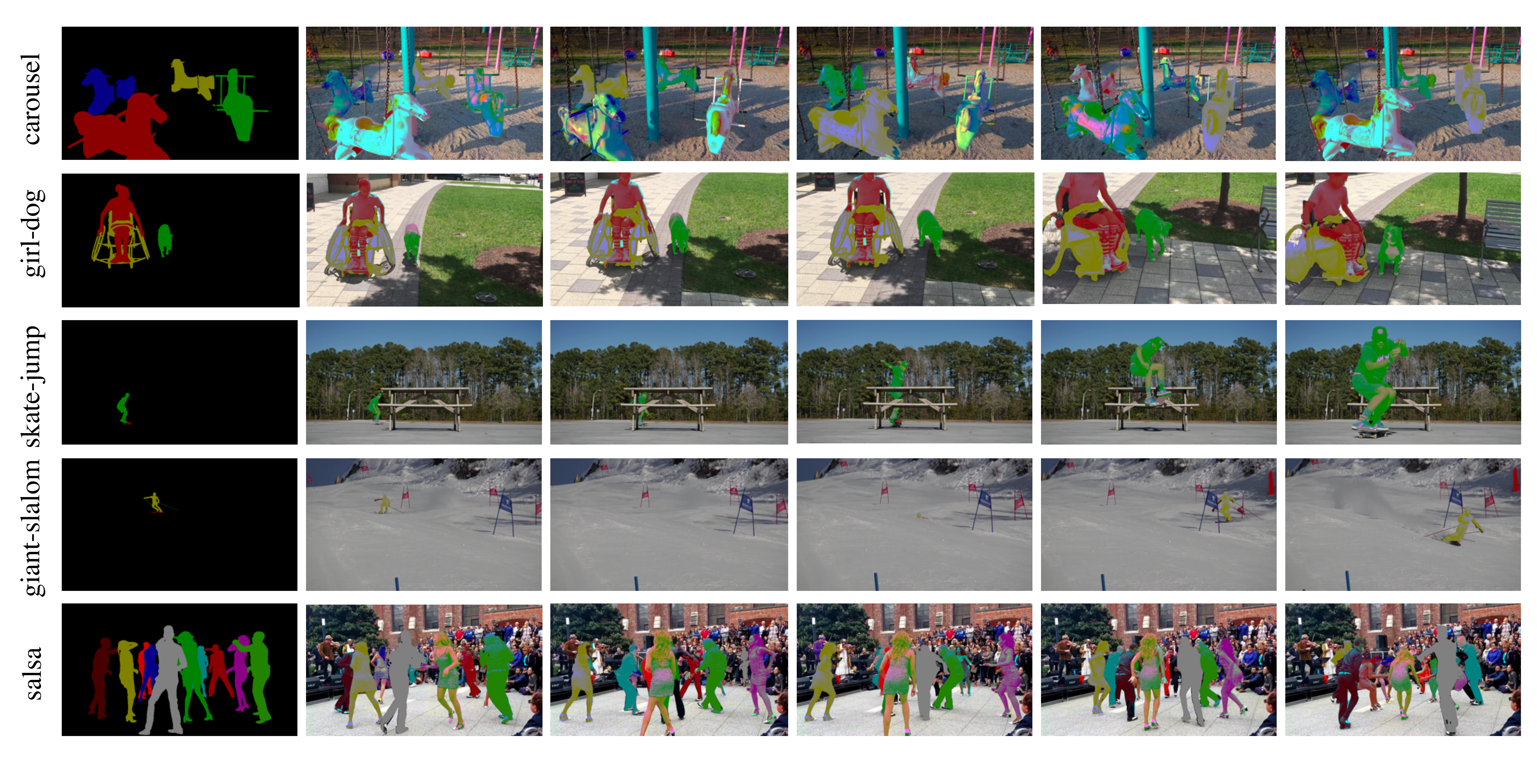}
\end{center}
   \caption{Visualization results of OVS-Net's prediction. The first column shows the first frame ground-truth masks of each video sequence.~The frames are chosen randomly. \textbf{Best viewed in color.}}
\label{fig:short}
\end{figure*}

\noindent \textbf{Effectiveness of the Inference Strategy} 

We conduct a series of comparative experiments to carefully choose $\rho$$_{iou}$. The $\rho$ $_{iou}$  is also important in IOU path as mentioned before. In our cascaded strategy, the object association is almost solved in IOU path.~Table \ref{iou_ab} demonstrates the final performance at different values. We find that the OVS-Net achieves the best performance when the $\rho$$_{iou}$ is 0.3.

\begin{table}[h]
\begin{center}
\setlength{\tabcolsep}{1mm}{
\begin{tabular}{|c|c|c|c|c|}
\hline
Methods  & {\em J}-mean & {\em F}-mean & {\em G}-mean \\
\hline\hline
OVS-Net(with $\rho$$_{iou}$ = 0.1) & 62.9 & 64.8 & 63.9  \\
OVS-Net(with $\rho$$_{iou}$ = 0.2) & 61.3 & 67.4 & 64.4  \\
OVS-Net(with $\rho$$_{iou}$ = 0.3) & \textbf{62.5} & \textbf{68.4} & \textbf{65.5}  \\
OVS-Net(with $\rho$$_{iou}$ = 0.4) & 61.7 & 62.9 & 62.3 \\
\hline
\end{tabular}}
\end{center}
\caption{Ablation study for  Inference Strategy with DAVIS$_{17}$ test-dev dataset. $\rho$$_{iou}$ means the threshold for the IOU mentioned before.}
\label{iou_ab}
\end{table}

\noindent \textbf{Effectiveness of each module of our OVS-Net}

Finally, we conduct a ablation study to evaluate each module of OVS-Net with DAVIS$_{17}$ test-dev dataset. We add effective modules step by step and show their performance as shown in Table \ref{each_module}. All models in the experiments are end-to-end trained and only use the single scale (for DAVIS$_{17}$ test-dev, we fix the shorter edges of input images are 800) without muti-scale and horizontal flip testing in testing phase.~OVS-Net with Re-ID module outperforms baseline OVS-Net by 6.1 points. To further introduce the MGA module into our framework, it can also brings another 4.5 points performance.

\begin{table}[h]
\begin{center}
\setlength{\tabcolsep}{1mm}{
\begin{tabular}{|c|c|c|c|c|}
\hline
Methods  & {\em G}-mean &  $\Delta$ {\em G}-mean \\
\hline\hline
OVS-Net(base) & 54.9 & - \\
+ Re-ID Module & 61.0 & +6.1 \\
+MGA Module& 65.5 & +4.5 \\
\hline
\end{tabular}}
\end{center}
\caption{Ablation study for  each module of OVS-Net with DAVIS$_{17}$ test-dev dataset.}
\label{each_module}
\end{table}

\subsection{Speed Analysis}

In this section, we compare the speed of our model with the state-of-the-art methods.~Our model is the first unified detection-based pipeline without optical flow during the training phase, significantly reducing training time and training complexity. As far as we know, the online training is extremely time-cousuming with post-processing including Data Dreaming\cite{khoreva2017lucid} and per-dataset finetuning.~The full OnAVOS~\cite{conf/bmvc/VoigtlaenderL17} takes roughly 13 seconds per frame and achieves 85.7 mIOU on DAVIS$_{16}$ val dataset.~The VS-ReID~\cite{xiaoxiao2017video}'s speed is about 0.33 FPS and the DyeNet~\cite{xiaoxiao-18video} is quicker and their speed is about 2.4 FPS without per-dataset finetuning, after per-dataset finetuning, their runing time is 0.43 FPS. According to the above running time, these approaches are far from the real-time applications.~Identical to the state-of-the-art DyeNet work \cite{conf/eccv/LiL18}, we use a single TitanXP GPU hardware and pytorch to perform evaluations for speed analysis throughout all the experiments. Following \cite{premvos2018}, for methods that only publish runtime results on the DAVIS$_{17}$ dataset, we take these timings as per object timings and extrapolate to the number of objects in the DAVIS$_{16}$ val dataset. Moreover, we directly quote the results of time consumption and accuracy of other algorithms mentioned in \cite{conf/eccv/LiL18} for fair comparison. As shown in Table \ref{speed}, our offline OVS-Net achieves the 83.1 mIOU on DAVIS$_{16}$ and the running time is \textbf{11.5 FPS}.~After online per-dataset training, our OVS-Net achieves 84.6 mIoU, and the corresponding running time is \textbf{2.3 FPS}.

\begin{table}[h]
\begin{center}
	\scalebox{0.8}{
\begin{tabular}{|c|c|c|c|}
\hline
Methods  & {\em G}-mean & speed/per frame & FPS\\
\hline\hline
OnAVOS~\cite{conf/bmvc/VoigtlaenderL17} & 85.7 & 13s & 0.08FPS \\
VS-ReID~\cite{xiaoxiao2017video} & - & 3s &  0.33FPS \\
DyeNet (OTS)~\cite{xiaoxiao-18video} & 84.7 & 0.42s & 2.4FPS \\
DyeNet~\cite{xiaoxiao-18video}$^{*}$ & 86.2 & 2.3s & 0.43FPS \\
PReMVOS (OTS)~\cite{premvos2018} & - & 0.5s & 2.0FPS \\
PReMVOS~\cite{premvos2018}$^{*}$ & 86.8 & 15.1s & 0.07FPS \\
\hline
OVS-Net (OTS) & 83.1 & \textbf{0.087s} & \textbf{11.5FPS} \\
OVS-Net $^{*}$ & 84.6 & \textbf{0.43s} & \textbf{2.3FPS} \\
\hline
\end{tabular}}
\end{center}
\caption{The inference speed analysis with DAVIS$_{16}$ val dataset.~* means after per-dataset finetuning using the first frame of testing dataset in their method. OTS means that the model is only trained on the training set without any annotations from the test set.}
\label{speed}
\end{table}

\section{Conclusion}

We propose the first unified detection-based pipeline for challenging video object segmentation, as well as a spatio-temporal training framework, which seamlessly integrates object detection, object segmentation, and object re-identification.~In principle, the proposed scheme is simple yet effective with a one-pass sequential inference strategy with online Re-ID, which ensures the computational efficiency of video object segmentation towards real-time performance (around \textbf{11.5 FPS}).~The mask-guided attention module is proposed to fully model the multi-scale object boundary information, which leads to performance improvements in object boundary.~Extensive experiments demonstrate the performance effectiveness and efficiency of this work against the state-of-the-art.

{\small
\bibliographystyle{ieee}
\bibliography{egbib}
}

\end{document}